\documentclass[sigconf]{acmart}
\usepackage{bm}
\usepackage{multirow}
\def\model{AITM} 
\def\module{AIT} 
\def\impression{$impression$}
\def\click{$click$}
\def\apply{$application$}
\def\credit{$approval$}
\def\activate{$activation$}
\def\purchase{$purchase$}
\def\constraint{Behavioral Expectation Calibrator}
\def\x{\bm{x}}

\def\v{\bm{v}}
\def\p{\bm{p}}
\def\q{\bm{q}}
\def\u{\bm{u}}

\def\z{\bm{z}}

\def\setR{\mathbb{R}}
\def\L{\mathcal{L}}
\def\D{\mathcal{D}} 

\AtBeginDocument{%
  \providecommand\BibTeX{{%
    \normalfont B\kern-0.5em{\scshape i\kern-0.25em b}\kern-0.8em\TeX}}}

\copyrightyear{2021} 
\acmYear{2021} 
\setcopyright{acmcopyright}\acmConference[KDD '21]{Proceedings of the 27th ACM SIGKDD Conference on Knowledge Discovery and Data Mining}{August 14--18, 2021}{Virtual Event, Singapore}
\acmBooktitle{Proceedings of the 27th ACM SIGKDD Conference on Knowledge Discovery and Data Mining (KDD '21), August 14--18, 2021, Virtual Event, Singapore}
\acmPrice{15.00}
\acmDOI{10.1145/3447548.3467071}
\acmISBN{978-1-4503-8332-5/21/08}

\begin{document}
\fancyhead{}

\title{Modeling the Sequential Dependence among Audience Multi-step Conversions with Multi-task Learning in Targeted Display Advertising}
\fancyhead{}

\author{
Dongbo Xi$^{1,*}$,
Zhen Chen$^{1,*}$,
Peng Yan$^{1}$,
Yinger Zhang$^{1,2}$,\\
Yongchun Zhu$^{3,4}$,
Fuzhen Zhuang$^{5,6}$, and
Yu Chen$^{1}$
}
\thanks{$*$ Corresponding authors: Dongbo Xi and Zhen Chen.}
\affiliation{
\institution{
$^1$Meituan\\
$^2$Zhejiang University, Hangzhou 310027, China \\
$^3$Key Lab of Intelligent Information Processing of Chinese Academy of Sciences (CAS),\\
Institute of Computing Technology, CAS, Beijing 100190, China\\
$^4$University of Chinese Academy of Sciences, Beijing 100049, China\\
$^5$Institute of Artificial Intelligence, Beihang University, Beijing 100191, China\\
$^6$SKLSDE, School of Computer Science, Beihang University, Beijing 100191, China\\
\{xidongbo,chenzhen06,yanpeng04,chenyu17\}@meituan.com,\\
zhangyinger@zju.edu.cn,
zhuyongchun18s@ict.ac.cn,
zhuangfuzhen@buaa.edu.cn
}
\country{}
}
\def\authors{Dongbo Xi, Zhen Chen, Peng Yan, Yinger Zhang, Yongchun Zhu, Fuzhen Zhuang, and Yu Chen}

\renewcommand{\shortauthors}{Xi and Chen, et al.}
\begin{abstract}
In most real-world large-scale online applications (e.g., e-commerce or finance), customer acquisition  is usually a multi-step conversion process of audiences. For example, an \impression~ $\rightarrow$ \click~ $\rightarrow$ \purchase~ process is usually performed of audiences for e-commerce platforms.
However, it is more difficult to acquire customers in financial advertising (e.g., credit card advertising) than in traditional advertising.
On the one hand, the audience multi-step conversion path is longer, an \impression~ $\rightarrow$ \click~ $\rightarrow$ \apply~ $\rightarrow$ \credit~ $\rightarrow$ \activate~ process usually occurs during the audience conversion for credit card business in financial advertising. 
On the other hand, the positive feedback is sparser (class imbalance) step by step, and it is difficult to obtain the final positive feedback due to the delayed feedback of \activate. 
Therefore, it is necessary to use the positive feedback information of the former step to alleviate the class imbalance of the latter step.
 Multi-task learning is a typical solution in this direction. While considerable multi-task efforts have been made in this direction, a long-standing challenge is how to explicitly model the long-path sequential dependence among audience multi-step conversions for improving the end-to-end conversion.
  In this paper, we propose an Adaptive Information Transfer Multi-task (\model) framework, which models the sequential dependence among audience multi-step conversions via the Adaptive Information Transfer (\module) module. The \module~ module can adaptively learn what and how much information to transfer for different conversion stages. Besides, by combining the \constraint~ in the loss function, the \model~framework can yield more accurate end-to-end conversion identification.
The proposed framework is deployed in Meituan app, which utilizes it to real-timely show a banner to the audience with a high end-to-end conversion rate for Meituan Co-Branded Credit Cards. 
  Offline experimental results on both industrial and public real-world datasets clearly demonstrate that the proposed framework achieves significantly better performance compared with state-of-the-art baselines.
  Besides, online experiments also demonstrate significant improvement compared with existing online models.
  Furthermore, we have released the source code of the proposed framework at https://github.com/xidongbo/AITM.
\end{abstract}




\maketitle

\section{Introduction}\label{introduction}
\begin{figure}[!t]
\begin{center}
\includegraphics[width=0.9\linewidth]{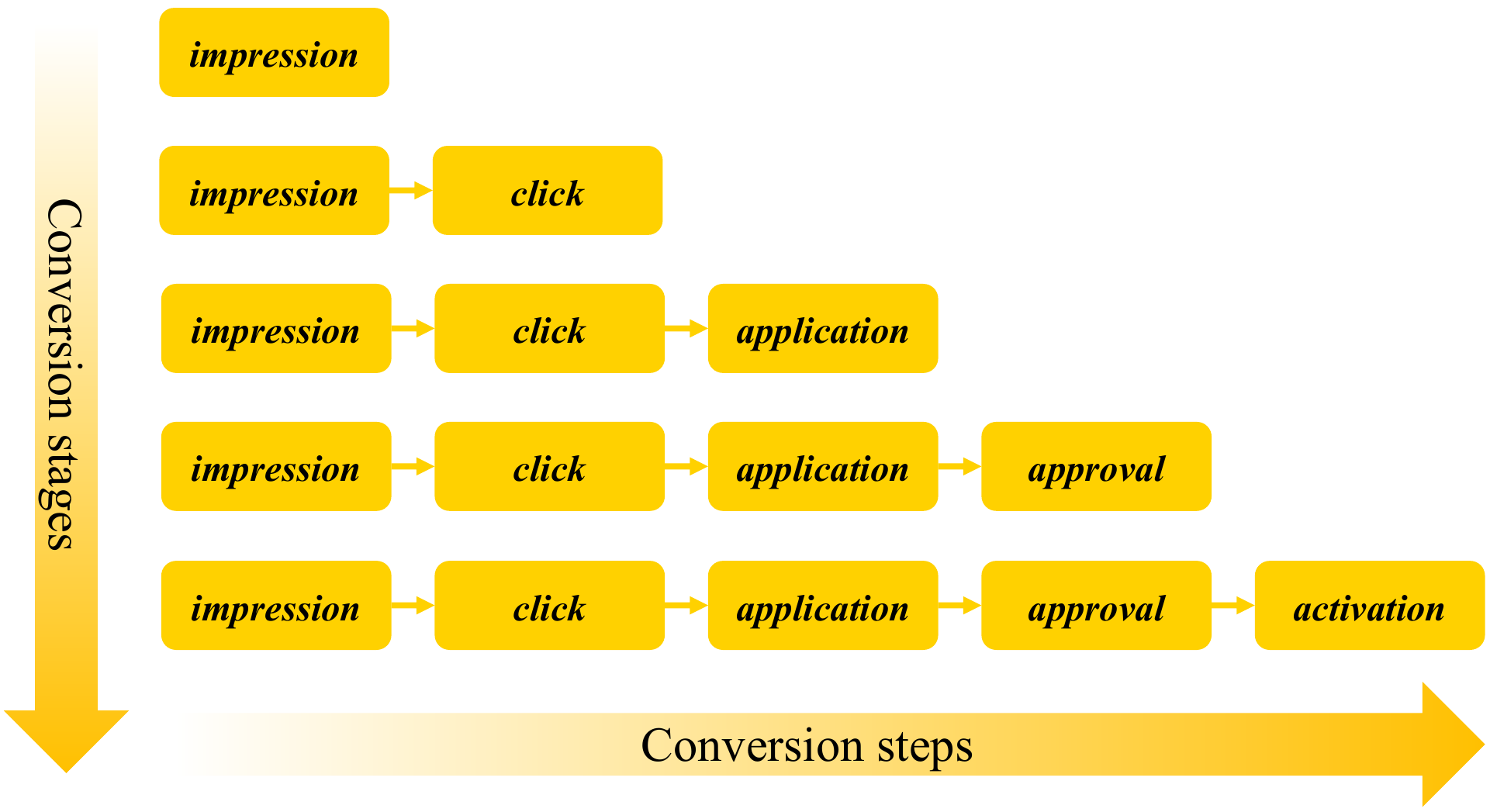}
\caption{There are five conversion steps with sequential dependence from left to right, and five different conversion stages of audiences from top to bottom. The lower the stage, the more efficient the conversion. In our business, we usually hope audiences to complete the last two stages.}
\label{fig:conversion_stages}%
\end{center}
\end{figure}%

Customer acquisition management can be considered the connection between advertising and customer relationship management to acquire new customers\footnote{https://en.wikipedia.org/wiki/Customer\_acquisition\_management}.
With the explosive growth of e-commerce, continuous and effective customer acquisition has become one of the biggest challenges for real-world large-scale online applications. 

In this paper, we focus on the customer acquisition task with sequential dependence among audience multi-step conversions.
Typically, in the credit card business, the audience multi-step conversion process usually needs to go through \impression~ $\rightarrow$ \click~ $\rightarrow$ \apply~ $\rightarrow$ \credit~ $\rightarrow$ \activate~ steps. These steps are defined as follows:

- \bm{\impression}: 
In our business, the \impression~
means that the advertising banner is shown to the audience selected according to several ranking metrics, e.g., the Click-Through Rate (CTR).

- \bm{\click}: The \click~ means that the shown banner is clicked by the audience, and redirected to the application page. 

- \bm{\apply}: The \apply~ means that the audience has filled in the application form and click the application button for a credit card.

- \bm{\credit}: The \credit~ means that the credit of the audience has been approved. In our system, this is also a real-time step.

- \bm{\activate}: The \activate~ is delayed feedback, and it means that the audience has activated the credit card within a period of time after \credit. Usually, we consider whether the audience has activated the credit card within 14 days (i.e., activation in T+14). 
The \activate~ feedback label is usually difficult to obtain due to the time-consuming of card mailing and the delayed feedback of audiences, so the class imbalance is more serious.

These conversion steps have sequential dependence, which means that only the former step occurs, the latter step may occur.
Based on this constraint, there are five different conversion stages of audiences as shown in Figure \ref{fig:conversion_stages}. Everything else is illegal. 

\begin{figure*}[!t]
\begin{center}
\includegraphics[width=0.9\linewidth]{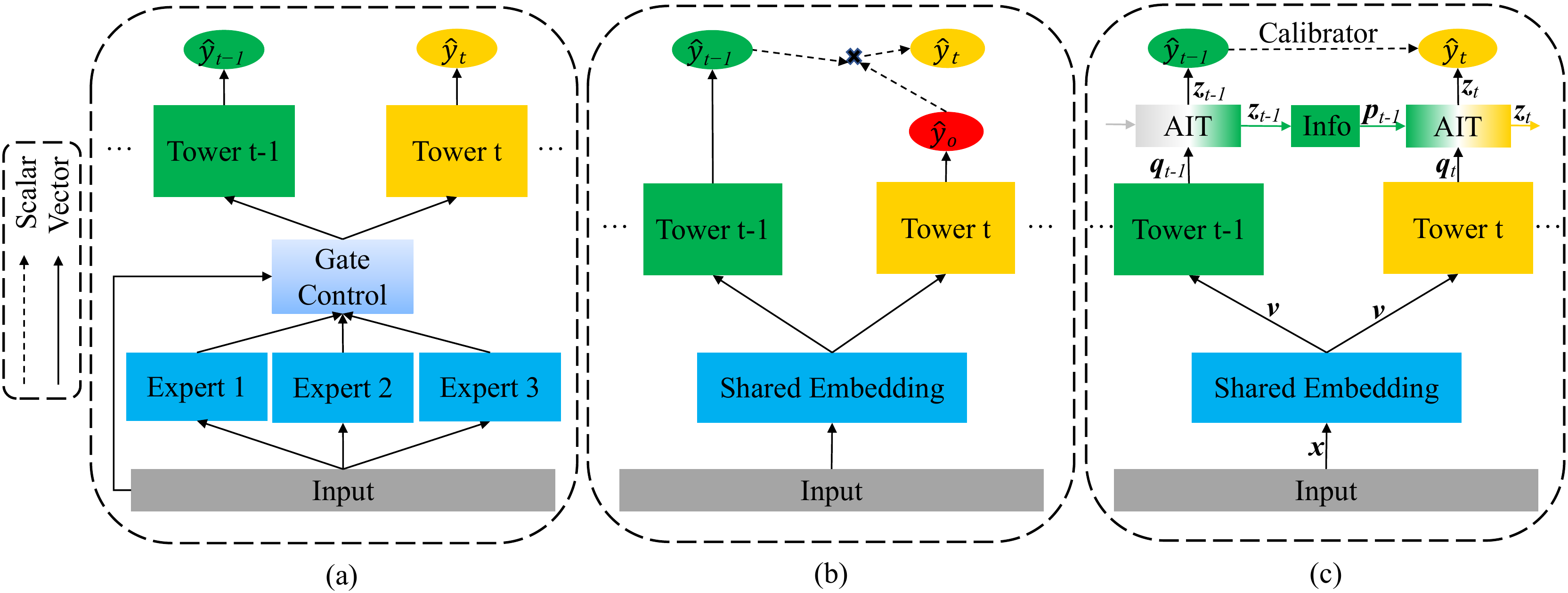}
\caption{(a) Expert-Bottom pattern. (b) Probability-Transfer pattern. The $\hat{y}_{o}$ is the non-end-to-end post-click conversion rate and the multi-task loss function only acts on the $\hat{y}_{t-1}$ and $\hat{y}_{t}$ in the original paper. (c) The proposed Adaptive Information Transfer Multi-task (\model) framework. For simplicity, only two adjacent tasks are shown in the figure.}
\label{fig:model}%
\end{center}
\end{figure*}

In industry and academia, multi-task learning is a typical solution to improve the end-to-end conversion in the audience multi-step conversion task.
Recently, considerable efforts have been done to model task relationships in multi-task learning. 
One idea is to control how Expert modules are shared across all tasks at the bottom of the multi-task model \cite{mmoe2018kdd,cgc2020recsys,mmoe2020seq}, and Tower modules at the top handle each task separately as shown in Figure \ref{fig:model} (a).
However, the Expert-Bottom pattern can only transfer shallow representations among tasks, but in the network close to the output layer, it often contains richer and more useful representations \cite{visual2014visualizing,visual2016convergent}, which have been proved to bring more gains \cite{mmd2014deep}.
Besides, the Expert-Bottom pattern is not specially designed for tasks with sequential dependence, so these models with the Expert-Bottom pattern can not model the sequential dependence explicitly.
Another idea is to transfer probabilities in the output layers of different tasks \cite{esmm2018sigir,npt2019neural,gao2019learning,esm22020sigir} as shown in Figure \ref{fig:model} (b). 
Similarly, the Probability-Transfer pattern can only transfer simple probability information via the scalar product, but richer and more useful representations are ignored in the vector space, which results in a great loss of gains.
If any one of the probabilities is not predicted accurately, multiple tasks will be affected.
Besides, the Probability-Transfer pattern is designed for solving the non-end-to-end post-click conversion rate via training on the entire space to relieve the sample selection bias problem, and these models with Probability-Transfer pattern can not model the sequential dependence well among audience multi-step conversions.
Therefore, a long-standing challenge is how to model the sequential dependence among audience multi-step conversions for improving the end-to-end conversion.

Along this line, we propose an Adaptive Information Transfer  Multi-task (\model) framework to model the sequential dependence among audience multi-step conversions. Specifically, due to the sequential dependence among audience multi-step conversions, the former conversion step (task) can bring useful information to the latter step (task). For example, if an audience has clicked the banner, then he/she may apply for the credit card. Conversely, if an audience doesn't click the banner, he/she certainly will not  apply for the credit card. 
Based on this, different conversion stages of different audiences need to transfer different information from the former step to the latter step, and as mentioned above, the vector space close to the output layer often contains richer and more useful information.
Therefore, we let the model adaptively transfer information in the vector space close to the output layer via the Adaptive Information Transfer (\module) module. 
Another advantage of the \module~module is that it can alleviate the class imbalance of the latter task with the help of the information from the former task, which has richer positive samples.
Also, because of the sequential dependence, the former task should have a higher end-to-end conversion probability than the latter for the same audience.
Therefore, we design a \constraint~ in the loss function.
On the one hand, it makes the model results more satisfy the real production constraint, on the other hand, it provides more accurate end-to-end conversion identification.
To summarize, the contributions of this paper are threefold:
\begin{itemize}
    \item The proposed \module~ module can adaptively learn what and how much information to transfer for different conversion stages of different audiences for improving the performance of multi-task learning with sequential dependence.
    \item Combining the \constraint~ in the loss function, offline experimental results on both industrial and public real-life datasets clearly demonstrate that the proposed framework achieves significantly better performance compared with state-of-the-art baselines.
    \item Online experiments also demonstrate significant improvement compared with existing online models, and the source code of the proposed framework has also been released.
\end{itemize}

\section{Related Work}
Multi-task learning (MTL) has led to successes in many applications of machine learning, from natural language processing and speech recognition to computer vision and drug discovery \cite{survey2017overview}.
In this section, we present the main multi-task learning works related to our work in two-fold:
the Expert-Bottom pattern and the Probability-Transfer pattern.

As shown in Figure \ref{fig:model} (a), the main idea of the Expert-Bottom pattern is to control how Expert modules are shared across all tasks at the bottom of the multi-task model \cite{mmoe2018kdd,cgc2020recsys,mmoe2020seq}, and the Tower modules at the top handle each task separately. Since complex problems may contain many sub-problems each requiring different experts \cite{moe22013learning}, some Mixture-of-Experts (MoE) models have been proposed one after another \cite{moe11991adaptive,moe22013learning,moe32017outrageously}. Inspired by the idea, \citeauthor{mmoe2018kdd} introduced the MoE to the multi-task learning and proposed the Multi-gate Mixture-of-Experts (MMoE) \cite{mmoe2018kdd} model by the gating networks assembling the experts for different tasks.  
\citeauthor{mmo22019video} explored a variety of soft-parameter sharing techniques such as MMoE to efficiently optimize for multiple ranking objectives for Video recommendation \cite{mmo22019video}.
\citeauthor{cgc2020recsys} proposed a Progressive Layered Extraction (PLE) \cite{cgc2020recsys} model to separate task-shared experts and task-specific experts explicitly. 
The Mixture of Sequential Experts (MoSE) model \cite{mmoe2020seq} has also been proposed to model sequential user behaviors in multi-task learning.  
However, the top Tower modules, which often contain richer and more useful information, can not help the tasks to improve each other due to there is no information exchange among them.

Another idea to model task relationships in multi-task learning is to transfer probabilities in the output layers of different tasks \cite{esmm2018sigir,npt2019neural,gao2019learning,esm22020sigir} as shown in Figure \ref{fig:model} (b).
\citeauthor{esmm2018sigir} proposed an Entire Space Multi-task Model (ESMM) \cite{esmm2018sigir} to transfer probabilities in the output layers by post-impression click-through rate (CTR) multiplying post-click conversion rate (CVR) equals post-impression click-through$\&$conversion rate (CTCVR). 
Further, more tasks are decomposed for probability transfer in $ESM^2$ \cite{esm22020sigir}.
The Neural Multi-Task Recommendation (NMTR) \cite{npt2019neural,gao2019learning} has also been proposed to extend the Neural Collaborative Filtering (NCF) \cite{ncf2017neural} to multi-task learning and relate the model prediction probability of each task in a cascaded manner.
However, as mentioned in Section \ref{introduction}, the Probability-Transfer pattern can only transfer simple probability information via the scalar product, but richer and more useful representations are ignored in the vector space, which results in a great loss. 
Besides, if any one of the probabilities is not predicted accurately, multiple tasks will be affected.

Other efforts have also utilized the tensor factorization \cite{tensor2017deep}, tensor normal priors \cite{tensor2017priors}, attention mechanism \cite{attention12019end,attention2019z}, and so on to solve the multi-task learning.
Nevertheless, these above efforts are not specially designed for tasks with sequential dependence, and they can not model the sequential dependence well among audience multi-step conversions.
\section{The MTL Ranking System in Meituan app}\label{sec:main_task}
\begin{figure*}[!t]
\begin{center}
\includegraphics[width=0.95\linewidth]{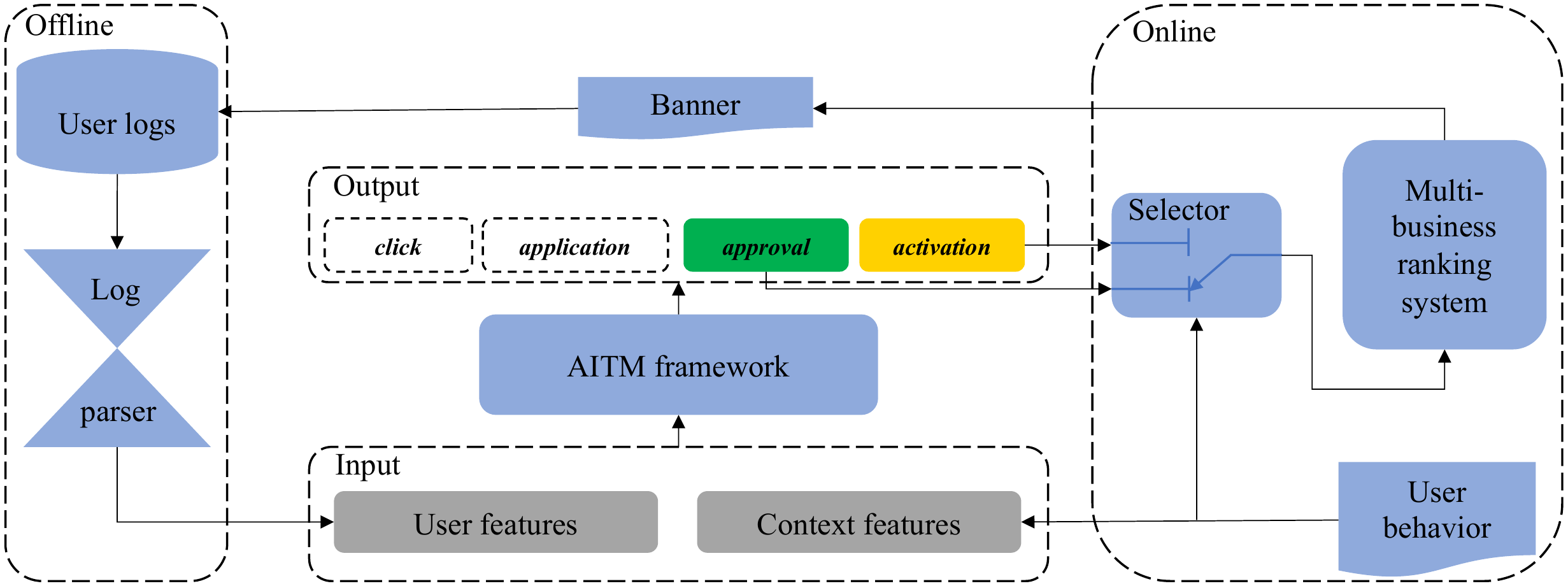}
\caption{The MTL Ranking System in Meituan app.}
\label{fig:system}%
\end{center}
\end{figure*}%
In this section, we give an overview of the MTL ranking system in Meituan app.
As shown in Figure \ref{fig:system}, in our credit card business, we model four tasks except for the passive \impression~step. 
Among them, the \credit~and \activate~are the main tasks, and the \click~and \apply~are the auxiliary tasks. 
That is because if the audience has only completed the \click~ and \apply~ steps, but the \credit~ step has not been completed, then it will cause a waste of resources (e.g., the computing and traffic resources). 
Because different audiences have different values to different businesses, the traffic that is useless to the credit card business may be useful to other businesses.
For this kind of audience, we might as well give the traffic to other businesses that may promote the audience conversion.
Therefore, we mainly focus on the last two end-to-end conversion tasks, i.e., \impression $\rightarrow$ \credit~and \impression $\rightarrow$ \activate.
Besides, because the last two tasks have fewer positive samples and the \activate~is delayed feedback, the first two auxiliary tasks with more positive samples can alleviate the class imbalance problem via the Adaptive Information Transfer module.

Meituan Co-Branded Credit Cards are issued in cooperation with different banks, and different banks are in different stages of business development, so they have different requirements for the \credit~and \activate. 
Start-up banks often want to issue more credit cards to quickly occupy the market, while mature banks want to increase the activation rate to achieve rapid profits.
Therefore, there is a selector in our system to output different conversion objectives for different banks.
The multi-task framework can deal with different business requirements well.

Besides, because different businesses in Meituan all need the traffic to acquire customers for their own business, and the sensitivities of different audiences to different businesses are different, so we can not simply divide the traffic into different businesses. 
We need a ranking mechanism to maximize the overall gain.
The multi-business ranking system ranks the different business scores according to the Equation:
\begin{eqnarray}
score = weight\times\hat{y},
\end{eqnarray}
where $\hat{y}$ is the predicted conversion probability for each business, and the $weight$ includes the value of the audience itself, the value of the business itself, and the value of the audience to the business.
The business banner with the highest score is shown to the audience.

\section{Methodology}
In this section, we first formulate the problem, then
we present the details of the proposed framework \model~ as shown in Figure \ref{fig:model} (c).

\subsection{Problem Formulation}
Given the input feature vector $\x$, assuming the audience needs $T$ steps to complete the final conversion after \impression~ (In Figure \ref{fig:conversion_stages}, $T=4$). In each conversion step $t$, if the audience completes the conversion step, the label $y_t$ is $1$, otherwise it is $0$. The sequential dependence means that $y_1\geq y_2\geq\cdots\geq y_T$ ($y_t\in \{0, 1\}, t = 1, 2, \cdots, T$). 
The multi-task framework needs to predict the end-to-end conversion probability $\hat{y}_t$ of each conversion step $t$ based on the input features $\x$:
\begin{equation}
\hat{y}_t = p(y_1=1,y_2=1,\cdots,y_t=1|\x).
\end{equation}%

\subsection{Adaptive Information Transfer Multi-task (\model) framework}
As shown in Figure \ref{fig:model} (c), given the input feature vector $\x$, 
we embed each entry $x_i$ ($x_i\in\x$, $1\leq i\leq|\x|$) to a low dimension dense vector representation $\v_i\in\setR^d$, where $d$ is the dimension of embedding vectors. The output of the Shared Embedding module is the concatenation of all embedding vectors: 
\begin{eqnarray}
\v = [\v_1;\v_2;\cdots;\v_{|\x|}],
\end{eqnarray}
where $[\cdot;\cdot]$ denotes the concatenation of two vectors.
By sharing the same embedding vectors among all tasks, on the one hand, the framework could learn the embedding vectors with rich positive samples of the former tasks to share information and alleviate the class imbalance of the latter tasks, and reduce the model parameters on the other hand.

Given $T$ tasks, the output of the Tower of each task $t (1\leq t\leq T)$ is defined as:
\begin{eqnarray}
\q_t=f_t(\v),
\end{eqnarray}
where the $f_t(\cdot)$ function is the Tower, $\q_t\in\setR^k$ and $k$ is the output dimension of the Tower. 
It should be mentioned that designing a different Tower is not the focus of this paper as we aim at designing an Adaptive Information Transfer module to model the sequential dependence.
In fact, our approach is a general framework, and any advanced models (e.g., NFM \cite{he2017nfm}, DeepFM \cite{guo2017deepfm}, AFM \cite{xiao2017afm}, and even the sequence models NHFM \cite{nhfm2020neural}, DIFM \cite{difm2021modeling}) can be easily integrated into our framework to act as the Tower, making the proposed \model~general and flexible.

For two adjacent tasks $t-1$ and $t$, the output of the AIT module of the task $t$ is computed as:
\begin{eqnarray}
\z_t&=&AIT(\p_{t-1}, \q_t), \\
where~\p_{t-1}&=&g_{t-1}(\z_{t-1}),
\end{eqnarray}
$\z_{t-1}\in\setR^k$ is the output of the AIT module of the task $t-1$, $g_{t-1}(\cdot)$ is the function to learn what information to transfer between the tasks $t-1$ and $t$, and $\p_{t-1}\in\setR^k$ is the learned transfer information.

The AIT module is designed to adaptively allocate the weights of the transferred information $\p_{t-1}$ and original information $\q_t$:
\begin{eqnarray} 
\z_t&=&\sum_{\u\in\{\p_{t-1},\q_t\}}w_uh_1(\u), \label{eqt:att}
\end{eqnarray}
where $w_u$ is the weight which is formulated as:
\begin{eqnarray} 
w_u=\frac{exp(\hat{w}_u)}{\sum_{u}exp(\hat{w}_u)},~~
\hat{w}_u=\frac{<h_2(\u),h_3(\u)>}{\sqrt{k}},
\label{eqt:weight}
\end{eqnarray}
where $<\cdot~, \cdot>$ represents the dot product.
$h_1(\cdot)$, $h_2(\cdot)$, and $h_3(\cdot)$ represent the feed-forward networks to project the input information to one new vector representation. There are lots of ways to design $h_1(\cdot)$, $h_2(\cdot)$, and $h_3(\cdot)$. In this paper, we use a simple single-layer MLP (Multi-Layer Perceptron) \cite{gardner1998mlp} as $h_1(\cdot)$, $h_2(\cdot)$, and $h_3(\cdot)$.
The idea of this kind of attention mechanism is similar to self-attention \cite{self2017attention}, the $h_1(\cdot)$, $h_2(\cdot)$ and, $h_3(\cdot)$ first learn Value, Query, Key from the same input $\u$, respectively. Then, we compute the similarity between Query ($h_2(\cdot)$) and Key ($h_3(\cdot)$) according to Equation (\ref{eqt:weight}). Finally, the Value ($h_1(\cdot)$) is weighted via the similarity according to Equation \ref{eqt:att}. 
This kind of attention mechanism has been proved to be more effective in the previous works \cite{nhfm2020neural,www2020modeling,difm2021modeling}. 

For the first task without the former task, the out of the AIT module is initialized to:
\begin{eqnarray} 
\z_1=\q_1.
\end{eqnarray}

The prediction probability of each task $t$ is:
\begin{eqnarray} 
\hat{y}_t=sigmoid(MLP(\z_t)),
\end{eqnarray}
where the MLP is used to project the $\z_t$ to the output space.

\subsection{\constraint~ and Joint Opratortimization for MTL}
For classification tasks, we need to minimize the \textit{cross-entropy} loss of all tasks:
\begin{equation}
\small{
\L_{ce}(\theta)=-\frac{1}{N}\sum_{t=1}^{T}\sum^N_{(\x,y_t)\in\D}(\left(y_t\log\hat{y}_t+(1-y_t)\log(1-\hat{y}_t)\right)},
\end{equation}
where $N$ is the number of samples in the entire sample space $\D$, $y_t$ is the label of the $t$-th task and $\theta$ is the parameter set in the MTL framework.

Besides, because of the sequential dependence, the former task should have a higher end-to-end conversion probability than the latter for the same audience, i.e., $\hat{y}_{t-1}\geq\hat{y}_t$.
We design a \constraint~ to minimize the following objective. On the one hand, it makes the model results more satisfy the real production constraint, on the other hand, it provides more accurate end-to-end conversion identification: 
\begin{equation}
\L_{lc}(\theta)=\frac{1}{N}\sum_{t=2}^{T}\sum^N_{\x\in\D}max(\hat{y}_t-\hat{y}_{t-1},0).
\end{equation}%
If $\hat{y}_t>\hat{y}_{t-1}$, the $\L_{lc}(\theta)$ will output a positive penalty term, otherwise output 0.

The final loss function $\L(\theta)$ of the \model~ combines the two components to a unified multi-task learning framework:
\begin{equation}
\L(\theta)=\L_{ce}(\theta)+\alpha\L_{lc}(\theta),
\end{equation}%
where $\alpha$ controls the strength of the \constraint~ component.

The framework is implemented using TensorFlow\footnote{https://www.tensorflow.org/} and trained through stochastic gradient descent over shuffled mini-batches with the Adam \cite{adam2015method} update rule.
\section{Experiments}
\begin{table}[!t]
  \centering
  \caption{Summary statistics for the datasets. ``\%Positive'' represents the percentage of positive samples in the train set over each task.}
    \setlength{\tabcolsep}{0.6mm}{
    \begin{tabular}{cccccc}
    \toprule
    Dataset & \multicolumn{1}{l}{\#Task} & \#Train & \#Validation & \#Test & \%Positive(\%) \\
    \midrule
    Industrial & 4     & 20M   & 3M    & 26M   & 23.29/1.84/1.30/1.00 \\
    Public & 2     & 38M   & 4.2M  & 43M   & 3.89/0.02 \\
    \bottomrule
    \end{tabular}
    }%
  \label{tab:dataset}%
\end{table}%
In this section, we perform experiments to evaluate the proposed framework against various baselines on both industrial and public real-world datasets.
We first introduce the datasets, evaluation protocol, and baseline methods. Finally, we present our experimental results and analysis.
\begin{table*}[!t]
  \centering
  \caption{The AUC performance (mean$\pm$std) on the industrial and public datasets. The Gain means the mean AUC improvement compared with the LightGBM. Underlined results indicate the best baselines over each task. ``*'' indicates that the improvement of the proposed \model~ is statistically significant compared with the best baselines at p-value $<$ 0.05 over paired samples t-test, and ``**'' indicates that the p-value $<$ 0.01.}
    \begin{tabular}{c|cccc|cccc}
    \toprule
    \multirow{2}[2]{*}{Model} & \multicolumn{4}{c|}{Industrial dataset} & \multicolumn{4}{c}{Public dataset} \\
          & \credit~ AUC & \activate~ AUC & \multicolumn{2}{c|}{Gain} & \click~ AUC & \purchase~ AUC & \multicolumn{2}{c}{Gain} \\
    \midrule
    LightGBM & 0.8392$\pm$0.0011 & 0.8536$\pm$0.0035 & -     & -     & 0.5837$\pm$0.0005 & 0.5870$\pm$0.0038 & -     & - \\
    MLP   & 0.8410$\pm$0.0010 & 0.8602$\pm$0.0014 & +0.0018  & +0.0066  & 0.6048$\pm$0.0013 & 0.5806$\pm$0.0035 & +0.0211  & -0.0064  \\
    \midrule
    ESMM  & 0.8443$\pm$0.0028 & 0.8691$\pm$0.0025 & +0.0051  & +0.0155  & 0.6022$\pm$0.0020 & 0.6291$\pm$0.0061 & +0.0185  & +0.0421  \\
    OMoE  & 0.8438$\pm$0.0022 & 0.8714$\pm$0.0009 & +0.0046  & +0.0178  & \textbf{\underline{0.6049$\pm$0.0020}} & 0.6405$\pm$0.0041 & \textbf{+0.0212} & +0.0535  \\
    MMoE  & 0.8444$\pm$0.0026 & 0.8705$\pm$0.0009 & +0.0052  & +0.0169  & 0.6047$\pm$0.0017 & \underline{0.6420$\pm$0.0031} & +0.0210  & +0.0550  \\
    PLE   & \underline{0.8518$\pm$0.0006} & \underline{0.8731$\pm$0.0016} & +0.0126  & +0.0195  & 0.6039$\pm$0.0014 & 0.6417$\pm$0.0013 & +0.0202  & +0.0547  \\
    AITM  & \textbf{0.8534$\pm$0.0011**} & \textbf{0.8770$\pm$0.0005*} & \textbf{+0.0142} & \textbf{+0.0234} & 0.6043$\pm$0.0016 & \textbf{0.6525$\pm$0.0024**} & +0.0206  & \textbf{+0.0655} \\
    \bottomrule
    \end{tabular}%
  \label{tab:result12}%
\end{table*}%

\subsection{Datasets}
\begin{itemize}
    \item \textbf{Industrial dataset:} The industrial dataset contains all samples that are shown a banner of Meituan Co-Branded Credit Cards over a continuous period of time. 
    We divide the training, validation, and test sets in chronological order. We downsample the \activate~ negative samples for each bank to keep the proportion of positive samples to be $1\%$ overall except for the test set. 
    Because it is necessary to evaluate the performance of the model on the test set that meets the real data distribution.
    Four tasks (i.e., \click, \apply, \credit, \activate) are contained in the dataset.
    \item \textbf{Public dataset:} The public dataset is the Ali-CCP (Alibaba Click and Conversion Prediction) \cite{esmm2018sigir} dataset\footnote{https://tianchi.aliyun.com/datalab/dataSet.html?dataId=408}. We use all the single-valued categorical features. Two tasks (i.e., \click, \purchase) are contained in the dataset. We randomly take $10\%$ of the train set as the validation set to verify the convergence of all models.
\end{itemize}
For these two datasets, we filter the features whose frequency less than 10.
The statistics of these datasets are shown in Table \ref{tab:dataset}.
\subsection{Evaluation Protocol}
In the offline experiments, to evaluate the performance of the proposed \model~framework and the baselines, we follow the existing works \cite{mmoe2018kdd,esmm2018sigir,esm22020sigir,cgc2020recsys} to use the standard metric: \textbf{AUC} (Area Under ROC).
In ranking tasks, AUC is a widely used metric to evaluate the ranking ability. The mean and  standard deviation (std) are reported over five runs with different random seeds.
In the online A/B test, we use the \textbf{end-to-end conversion rate} to evaluate the performance more intuitively.
On all datasets, we report the AUC of end-to-end tasks, which are directly optimized in their loss functions.
Besides, we only report the metrics on the focused main tasks (i.e., the \credit~and \activate~tasks) over the industrial dataset.

\subsection{Baseline Methods}
We compare the proposed method with the following competitive and mainstream models:
\begin{itemize}
    \item \textbf{LightGBM} \cite{lgb2017lightgbm}: LightGBM is a gradient boosting framework that uses tree based learning algorithms. LightGBM is being widely-used in many winning solutions of machine learning competitions\footnote{https://github.com/microsoft/LightGBM}.
    \item \textbf{MLP} \cite{gardner1998mlp}: We use the base structure of our \model~framework as the single task model. It is a Multi-Layer Perceptron.
    \item \textbf{ESMM} \cite{esmm2018sigir,esm22020sigir}: The ESMM and $ESM^2$ with Probability-Transfer pattern are designed for solving the non-end-to-end post-click conversion rate via training on the entire space to relieve the sample selection bias problem.
    \item \textbf{OMoE} \cite{mmoe2018kdd}: The OMoE with Expert-Bottom pattern integrates experts via sharing one gate among all tasks.
    \item \textbf{MMoE} \cite{mmoe2018kdd}: The MMoE with Expert-Bottom pattern is designed to integrate experts via multiple gates in the Gate Control as shown in Figure \ref{fig:model} (a).
    \item \textbf{PLE} \cite{cgc2020recsys}: The Progressive Layered Extraction (PLE) with Expert-Bottom pattern separates task-shared experts and task-specific experts explicitly. This is the state-of-the-art method under different task correlations.
\end{itemize}

\subsection{Performance Comparison}
\subsubsection{\textbf{Offline Results}}
\begin{figure*}[!t]
\begin{center}
\includegraphics[width=0.925\linewidth]{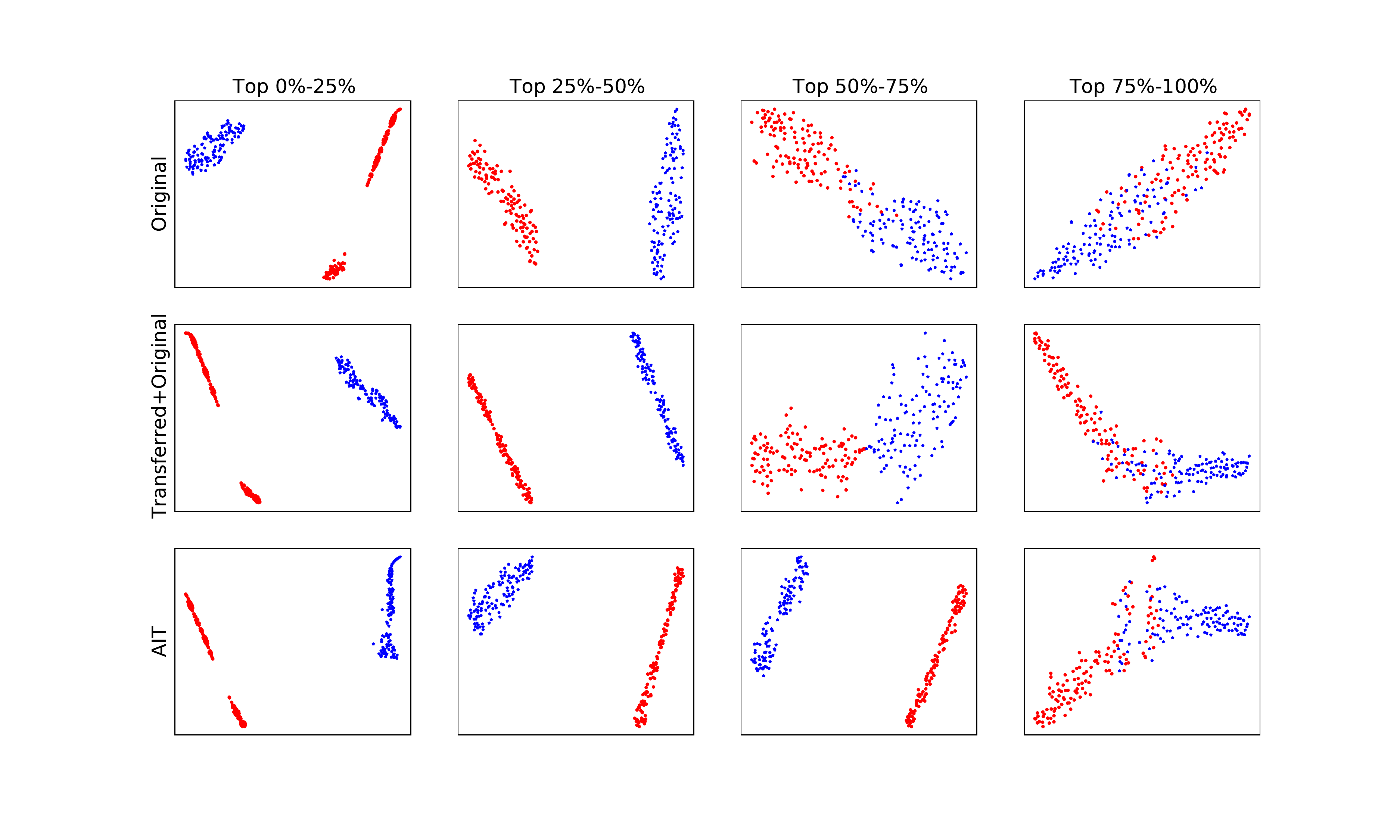}
\caption{The t-SNE visualization at different conversion score rankings of the original information $\q_t$, transferred plus original information  $\p_{t-1}+\q_t$  and the information $\z_t$ learned by the \module~ on the \textit{\textbf{activation}} task.}
\label{fig:activate}%
\end{center}
\end{figure*}%

In this subsection, we report the AUC scores and gains of all models on the offline test set. 
As mentioned in Section \ref{sec:main_task}, we only focus on the last two main end-to-end conversion tasks on the industrial dataset. 
The results of \credit~ AUC and \activate~ AUC are shown in Table \ref{tab:result12}.
From these results, we have the following insightful observations:
\begin{itemize}
    \item The MLP obtains $0.0018$ and $0.0066$ AUC gains on two tasks, respectively, compared with the tree-based model LightGBM, which indicates the fitting ability of neural network models on large-scale datasets.
    \item Compared with the single-task models LightGBM and MLP, the multi-task models ESMM, OMoE, MMoE, PLE and AITM obtain more gains by introducing the multi-task information in the neural networks.
    \item The Probability-Transfer pattern-based ESMM achieves a relatively small improvement due to only simple probability information is transferred between adjacent tasks.
    \item The Expert-Bottom pattern-based models obtain further performance improvement by controlling the shared information among different tasks. However, neither of the one-gate and multi-gate models is a clear winner on this dataset.
    \item The PLE obtains the best performance among these baselines on the two tasks via separating task-shared experts and task-specific experts explicitly.
    \item Our \model~achieves significant improvement compared with various state-of-the-art baseline models, which shows the \module~module is effective and could bring more gains on sequential dependence tasks.
\end{itemize}
The results on the public dataset are also shown in Table \ref{tab:result12}.
From these results, we have the following  findings, which are not the same as above:
\begin{itemize}
    \item Serious class imbalance on the \purchase~ task (the proportion of positive samples is $0.02\%$ as shown in Table \ref{tab:dataset}) leads to the poor performance of the two single-task models, i.e., the LightGBM and MLP.
    \item The MLP obtains similar performance improvement compared with multi-task models on the \click~ task. In other words, the multi-task models seem to have no significant improvement on the \click~ task. This may be because there are only two tasks in this dataset, and no other task can provide more information before the \click~ task. $3.89\%$ positive samples in the \click~ task are relatively abundant as shown in Table \ref{tab:dataset}.
    \item The Expert-Bottom pattern shows better performance than the Probability-Transfer pattern on the \purchase~ task with serious class imbalance. Besides, our \model~ can explicitly use the rich positive sample information of the former \click~ task to alleviate the class imbalance of the current \purchase~ task and achieve the best performance. On the other hand, it also shows the generalization ability of the proposed \model.
\end{itemize}

\subsubsection{\textbf{Online Results}}

\begin{table}[!t]
  \centering
  \caption{Online A/B test results.}
    \setlength{\tabcolsep}{5mm}{\begin{tabular}{ccc}
    \toprule
    \multirow{2}[2]{*}{Model} & \multicolumn{2}{c}{Gain} \\
          & \credit~ & \activate~ \\
    \midrule
    MLP vs LightGBM & +16.95\% & +17.55\% \\
    AITM vs MLP & +25.00\% & +42.11\% \\
    \bottomrule
    \end{tabular}
    }%
  \label{tab:result_online}%
\end{table}%

The proposed framework is trained offline and regularly updated. 
The pre-trained model is deployed in Meituan app by the TF Serving\footnote{https://github.com/tensorflow/serving}, to real-timely show a banner to the audience with a high end-to-end \credit~ or \activate~ conversion rate for Meituan Co-Branded Credit Cards.
Due to business competition, user experience, and delayed feedback (T+14) of the \activate~ task, we can not deploy all models to the online system. 
With the development of our business, we have successively deployed LightGBM, MLP, and AITM to the online system.
These models serve tens of millions of traffic every day.
A/B test is carried out for every two models with the same traffic for two consecutive weeks (It takes four weeks for all feedback to be received for every two models).
The online A/B test results are shown in Table \ref{tab:result_online}.
Compared with LightGBM, the \impression~ $\rightarrow$ \credit~ conversion rate of MLP increases by $16.95\%$ and the \impression~ $\rightarrow$ \activate~ conversion rate increases by $17.55\%$.
The \model~ further increases the \impression~$\rightarrow$\credit~ and \impression~ $\rightarrow$ \activate~ 
conversion rate by $25.00\%$, $42.11\%$ compared with MLP, respectively. 
Now, the \model~ has provided real-time prediction for all traffic in our system.

Besides, our system is computationally efficient, and the TP999, TP9999 of the real-time prediction is less than 20ms, 30ms in the system every day, respectively, which can meet the requirement of real-time solutions.

\begin{figure*}[!t]
\begin{center}
\includegraphics[width=0.95\linewidth]{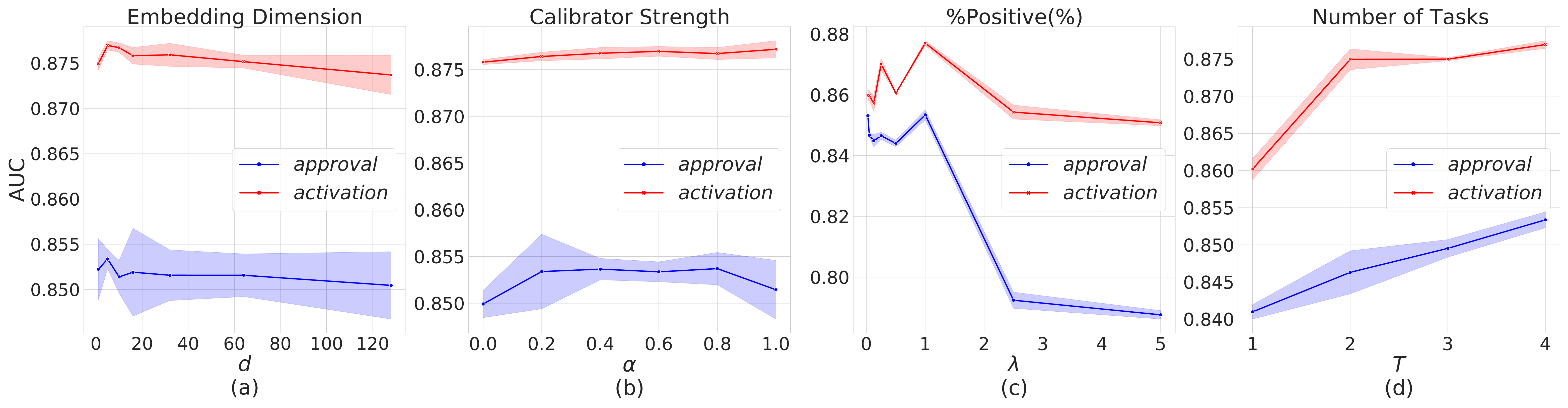}
\caption{The mean AUC performance in different experiment settings, the shaded part represents the standard deviation. (a) The impact of the embedding dimension $d$. (b) The impact of the strength $\alpha$ of the \constraint. (c) The impact of the proportion $\lambda$ of positive samples. (d) Ablation Study of the number $T$ of tasks.}
\label{fig:params}%
\end{center}
\end{figure*}%
\begin{figure}[!t]
\begin{center}
\includegraphics[width=0.95\linewidth]{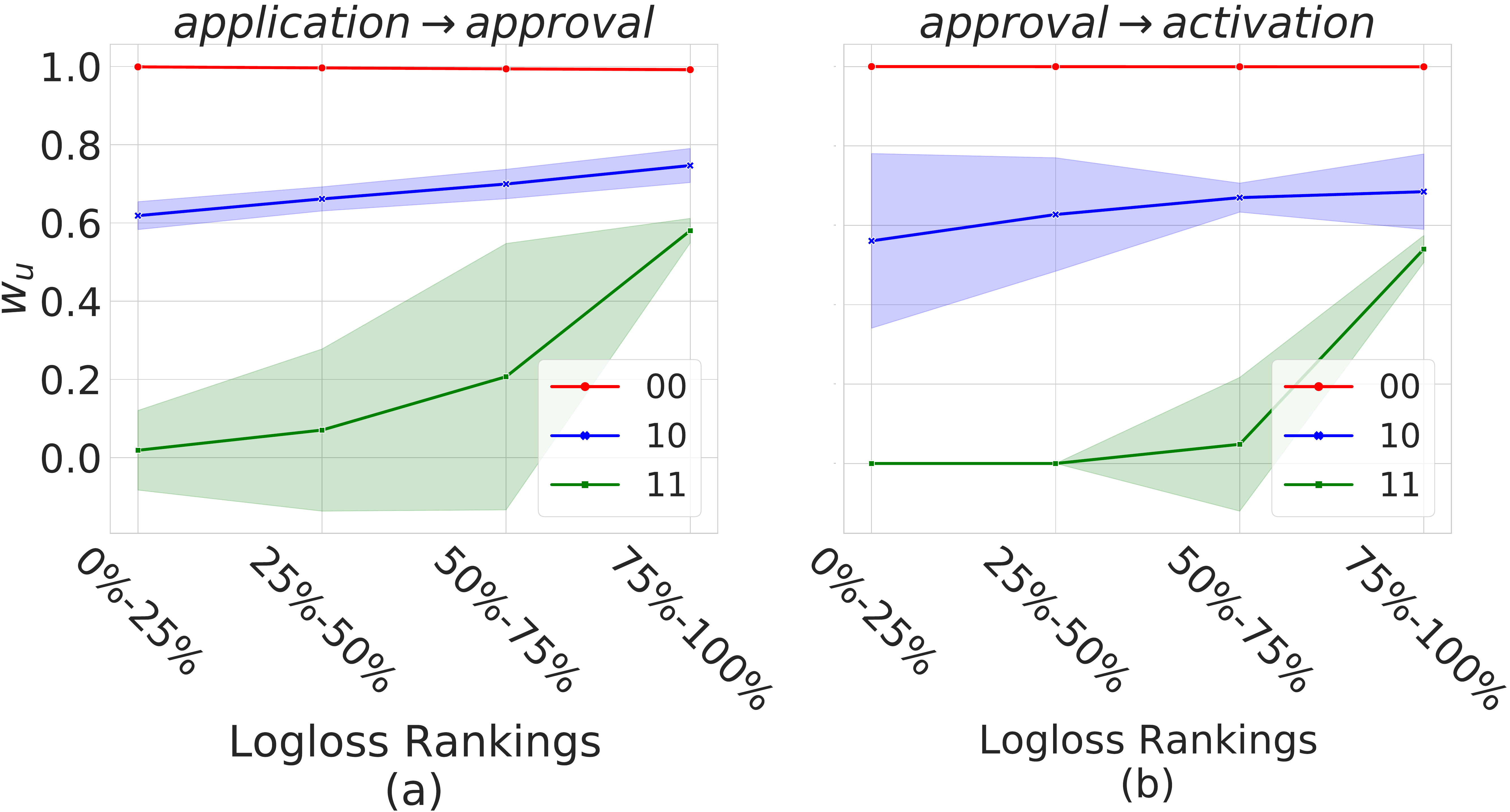}
\caption{The mean weight $w_u$ of the transferred information $\p_{t-1}$ over different conversion stages (00/10/11) at different logloss rankings, the shaded part represents the standard deviation. (a) The tasks \textit{\textbf{application}}~$\rightarrow$~\textit{\textbf{approval}}. (b) The tasks \textit{\textbf{approval}~}$\rightarrow$~\textit{\textbf{activation}}.}
\label{fig:weight}%
\end{center}
\end{figure}%
\subsection{Ablation Study}\label{sub:ablation}

In this subsection, we perform the ablation study of the \module~ module and the number of the tasks.

Firstly, we randomly sample $500$ \activate~ positive and negative samples in the test set, respectively.
The \activate~ prediction scores of positive samples are ranked in descending order, while those of negative samples are in ascending order.
We plot the original information $\q_t$, transferred plus original information $\p_{t-1}+\q_t$  and the information $\z_t$ learned by the \module~ on the \activate~ task in Figure \ref{fig:activate} via the t-SNE (t-distributed Stochastic Neighbor Embedding \cite{tsne2008}).
From the visualization, we can obtain the following inspiring observations (Similar results can also be observed in the \credit~ task, we list it in the appendix due to the space limitation):
\begin{itemize}
    \item When the prediction scores of the \model~ are very confident (see the Top $0\%-50\%$ in Figure \ref{fig:activate}), the three components (i.e., the Original, Transferred+Original, and AIT) can accurately identify positive and negative samples.
    \item With the confidence of the prediction scores of the \model~ decreases (see the Top $50\%-100\%$ in Figure \ref{fig:activate}), it is difficult to identify positive and negative samples only using the original information. The transferred plus original information improves the performance compared with only the original information, which indicates that information transfer could improve the performance of tasks with sequential dependence in multi-task learning.
    \item The \module~ module could adaptively learn what and how much information to transfer among audience multi-step conversions via the multi-task framework,
    so the \module~ further improves the performance compared with the transferred plus original information under low confidence.
\end{itemize}
Besides, we study the impact of the number of tasks as shown in Figure \ref{fig:params} (d). We perform the experiments over tasks \activate~ (\credit), \credit~ $\rightarrow$ \activate, \apply~ $\rightarrow$ \credit~ $\rightarrow$ \activate, $click~\\\rightarrow$ \apply~ $\rightarrow$ \credit~ $\rightarrow$ \activate, respectively. More tasks with more positive sample information and transferred information greatly improve the performance.

\subsection{Case Study}
In order to understand how much information the \module~ module transfers for different conversion stages, we extract the weight $w_u$ in Equation (\ref{eqt:weight}) of the transferred information $\p_{t-1}$ and show it in Figure \ref{fig:weight}.
We first randomly sample $40,000$ test samples.
In Figure \ref{fig:weight} (a), we divide the samples into three groups according to the \apply~ and \credit~ labels of $00/10/11$, and rank the top $500$ samples in each group according to the logloss of each sample in ascending order. Figure \ref{fig:weight} (b) is the same except for the tasks are \credit~ and \activate.
From Figure \ref{fig:weight}, we have the following interesting findings:
\begin{itemize}
    \item Because when the label of the former task is 0, the label of the latter task must also be 0, we can see that at this time the former task transfers very strong information to the latter task (the weight is close to 1 of the red lines in Figure \ref{fig:weight}).
    \item When the label of the former task is 1, the label of the latter task is uncertain. 
    When the label of the latter task is 1, there is little information is transferred from the former task (the green lines in Figure \ref{fig:weight}), which indicates that the latter task mainly identifies positive samples based on the task itself.
    \item When the label of the former task is 1, with the prediction becomes worse (the logloss rankings from 0\%-25\% to 75\%-100\%), the weight of the transferred information gradually increases (the blue and green lines in Figure \ref{fig:weight}), which indicates that the prediction result of the latter task is misled by the former task.
\end{itemize}
From the above results, we could see that the \module~ module can learn how much information to transfer between two adjacent tasks.

\subsection{Hype-parameter Study}
In order to study the impact of hype-parameters and the stability on the performance of the \model, we perform the hyper-parameter study.

Firstly, considering the embedding dimension $d$, we vary the embedding dimension as $[1,5,10,16,32,64,128]$, the results are shown in Figure \ref{fig:params} (a).
We can see that the performance of the \model~ is not very sensitive to the embedding dimension.
The embedding dimension is related to the complexity and capability of the model. Usually, smaller embedding dimension may fit the data distribution insufficiently, especially if the numbers of samples and features are large. While a larger embedding dimension increases the complexity of the model and requires more samples and features to fit, a proper embedding dimension can achieve the best performance \cite{bistddp2019modelling}. Making a trade-off between model complexity and capability, we finally set $d = 5$ as the embedding dimension in all experiments.

Secondly, we study the impact of the strength $\alpha$ of the \constraint~ as shown in Figure \ref{fig:params} (b). 
There are performance fluctuations (the seesaw phenomenon) among four different tasks. However, the \constraint~ brings the improvement of the overall performance. We finally set $\alpha = 0.6$ as the weight in all the experiments.

Thirdly, we study the impact of the proportion $\lambda$ of positive samples in the industrial dataset.
We downsample the \activate~ negative samples to keep the proportion $\lambda$ of positive samples to be $[0.025\%,0.05\%,0.125\%,0.25\%,$ $0.5\%,1\%,2.5\%,5\%]$ in the train set, respectively, and report the AUC performance on the entire test set in Figure \ref{fig:params} (c).
On the one hand, if audiences do not apply for the credit card at present, it does not mean that they will not apply for the card in the future, so we can not use too many negative samples for training.
On the other hand, it can be seen that when $\lambda$ is too large, the performance of the model drops sharply. This is because too much \activate~ negative sample information is lost.
Besides, excessive downsampling of the \activate~ negative samples also leads to the loss of \credit~ positive samples.
We finally downsample the \activate~ negative samples to keep the proportion $\lambda$ of positive samples to be $1\%$. 
This setting is applied to all models.

Combining the performance in Table \ref{tab:result12} and Figure \ref{fig:params}, we can see that even without the best parameters, the \model~ is still superior to other baselines in most cases. 
In other words, the performance of the \model~ stays stable in a large range of values of hyper-parameters and is not very sensitive to the hyper-parameters.
\section{Conclusion}
In this paper, we proposed an Adaptive Information Transfer Multi-task (\model) framework to model the sequential dependence among audience multi-step conversions. The proposed Adaptive Information Transfer (\module) module combining the \constraint~ in the loss function could learn what and how much information to transfer for different conversion stages for improving the performance of multi-task learning with sequential dependence.
Offline and online experimental results demonstrate significant improvement compared with state-of-the-art baseline models.


\section{Acknowledgments}
Fuzhen Zhuang is supported by the National Natural Science Foundation of China under Grant Nos. U1836206, U1811461.
Besides, we thank Zhenhua Zhang, Kun Chen, Chang Qu, Qiu Xiong, and KangMing Yu for their support and valuable suggestions.

\begin{figure*}[!t]
\begin{center}
\includegraphics[width=0.925\linewidth]{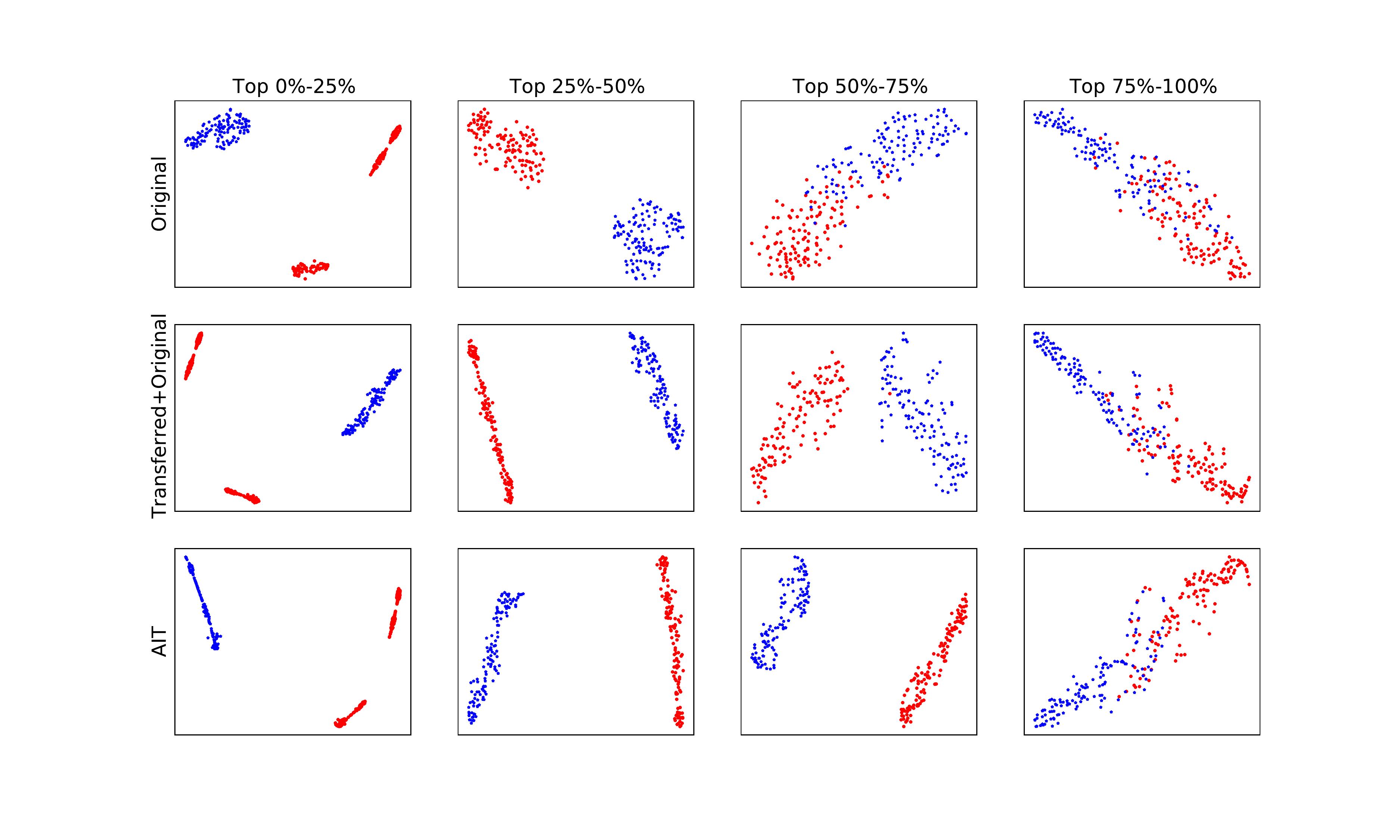}
\caption{The t-SNE visualization at different conversion score rankings of the original information $\q_t$, transferred plus original information  $\p_{t-1}+\q_t$  and the information $\z_t$ learned by the \module~ on the \textit{\textbf{approval}} task.}
\label{fig:credit}%
\end{center}
\end{figure*}%
\bibliographystyle{ACM-Reference-Format}
\bibliography{main}


\begin{thebibliography}{32}


\ifx \showCODEN    \undefined \def \showCODEN     #1{\unskip}     \fi
\ifx \showDOI      \undefined \def \showDOI       #1{#1}\fi
\ifx \showISBNx    \undefined \def \showISBNx     #1{\unskip}     \fi
\ifx \showISBNxiii \undefined \def \showISBNxiii  #1{\unskip}     \fi
\ifx \showISSN     \undefined \def \showISSN      #1{\unskip}     \fi
\ifx \showLCCN     \undefined \def \showLCCN      #1{\unskip}     \fi
\ifx \shownote     \undefined \def \shownote      #1{#1}          \fi
\ifx \showarticletitle \undefined \def \showarticletitle #1{#1}   \fi
\ifx \showURL      \undefined \def \showURL       {\relax}        \fi
\providecommand\bibfield[2]{#2}
\providecommand\bibinfo[2]{#2}
\providecommand\natexlab[1]{#1}
\providecommand\showeprint[2][]{arXiv:#2}

\bibitem[\protect\citeauthoryear{Eigen, Ranzato, and Sutskever}{Eigen
  et~al\mbox{.}}{2014}]%
        {moe22013learning}
\bibfield{author}{\bibinfo{person}{David Eigen}, \bibinfo{person}{Marc'Aurelio
  Ranzato}, {and} \bibinfo{person}{Ilya Sutskever}.}
  \bibinfo{year}{2014}\natexlab{}.
\newblock \showarticletitle{Learning factored representations in a deep mixture
  of experts}. In \bibinfo{booktitle}{\emph{ICLR}}.
\newblock


\bibitem[\protect\citeauthoryear{Gao, He, Gan, Chen, Feng, Li, Chua, and
  Jin}{Gao et~al\mbox{.}}{2019a}]%
        {npt2019neural}
\bibfield{author}{\bibinfo{person}{Chen Gao}, \bibinfo{person}{Xiangnan He},
  \bibinfo{person}{Dahua Gan}, \bibinfo{person}{Xiangning Chen},
  \bibinfo{person}{Fuli Feng}, \bibinfo{person}{Yong Li},
  \bibinfo{person}{Tat-Seng Chua}, {and} \bibinfo{person}{Depeng Jin}.}
  \bibinfo{year}{2019}\natexlab{a}.
\newblock \showarticletitle{Neural multi-task recommendation from
  multi-behavior data}. In \bibinfo{booktitle}{\emph{ICDE}}.
  \bibinfo{pages}{1554--1557}.
\newblock


\bibitem[\protect\citeauthoryear{Gao, He, Gan, Chen, Feng, Li, Chua, Yao, Song,
  and Jin}{Gao et~al\mbox{.}}{2019b}]%
        {gao2019learning}
\bibfield{author}{\bibinfo{person}{Chen Gao}, \bibinfo{person}{Xiangnan He},
  \bibinfo{person}{Danhua Gan}, \bibinfo{person}{Xiangning Chen},
  \bibinfo{person}{Fuli Feng}, \bibinfo{person}{Yong Li},
  \bibinfo{person}{Tat-Seng Chua}, \bibinfo{person}{Lina Yao},
  \bibinfo{person}{Yang Song}, {and} \bibinfo{person}{Depeng Jin}.}
  \bibinfo{year}{2019}\natexlab{b}.
\newblock \showarticletitle{Learning to Recommend with Multiple Cascading
  Behaviors}.
\newblock \bibinfo{journal}{\emph{TKDE}} (\bibinfo{year}{2019}).
\newblock


\bibitem[\protect\citeauthoryear{Gardner and Dorling}{Gardner and
  Dorling}{1998}]%
        {gardner1998mlp}
\bibfield{author}{\bibinfo{person}{Matt~W Gardner} {and} \bibinfo{person}{SR
  Dorling}.} \bibinfo{year}{1998}\natexlab{}.
\newblock \showarticletitle{Artificial neural networks (the multilayer
  perceptron)—a review of applications in the atmospheric sciences}.
\newblock \bibinfo{journal}{\emph{Atmospheric environment}}
  (\bibinfo{year}{1998}).
\newblock


\bibitem[\protect\citeauthoryear{Guo, Tang, Ye, Li, and He}{Guo
  et~al\mbox{.}}{2017}]%
        {guo2017deepfm}
\bibfield{author}{\bibinfo{person}{Huifeng Guo}, \bibinfo{person}{Ruiming
  Tang}, \bibinfo{person}{Yunming Ye}, \bibinfo{person}{Zhenguo Li}, {and}
  \bibinfo{person}{Xiuqiang He}.} \bibinfo{year}{2017}\natexlab{}.
\newblock \showarticletitle{DeepFM: a factorization-machine based neural
  network for CTR prediction}. In \bibinfo{booktitle}{\emph{IJCAI}}.
\newblock


\bibitem[\protect\citeauthoryear{He and Chua}{He and Chua}{2017}]%
        {he2017nfm}
\bibfield{author}{\bibinfo{person}{Xiangnan He} {and} \bibinfo{person}{Tat-Seng
  Chua}.} \bibinfo{year}{2017}\natexlab{}.
\newblock \showarticletitle{Neural factorization machines for sparse predictive
  analytics}. In \bibinfo{booktitle}{\emph{SIGIR}}.
\newblock


\bibitem[\protect\citeauthoryear{He, Liao, Zhang, Nie, Hu, and Chua}{He
  et~al\mbox{.}}{2017}]%
        {ncf2017neural}
\bibfield{author}{\bibinfo{person}{Xiangnan He}, \bibinfo{person}{Lizi Liao},
  \bibinfo{person}{Hanwang Zhang}, \bibinfo{person}{Liqiang Nie},
  \bibinfo{person}{Xia Hu}, {and} \bibinfo{person}{Tat-Seng Chua}.}
  \bibinfo{year}{2017}\natexlab{}.
\newblock \showarticletitle{Neural collaborative filtering}. In
  \bibinfo{booktitle}{\emph{TheWebConf}}. \bibinfo{pages}{173--182}.
\newblock


\bibitem[\protect\citeauthoryear{Jacobs, Jordan, Nowlan, and Hinton}{Jacobs
  et~al\mbox{.}}{1991}]%
        {moe11991adaptive}
\bibfield{author}{\bibinfo{person}{Robert~A Jacobs}, \bibinfo{person}{Michael~I
  Jordan}, \bibinfo{person}{Steven~J Nowlan}, {and} \bibinfo{person}{Geoffrey~E
  Hinton}.} \bibinfo{year}{1991}\natexlab{}.
\newblock \showarticletitle{Adaptive mixtures of local experts}.
\newblock \bibinfo{journal}{\emph{Neural computation}} \bibinfo{volume}{3},
  \bibinfo{number}{1} (\bibinfo{year}{1991}), \bibinfo{pages}{79--87}.
\newblock


\bibitem[\protect\citeauthoryear{Ke, Meng, Finley, Wang, Chen, Ma, Ye, and
  Liu}{Ke et~al\mbox{.}}{2017}]%
        {lgb2017lightgbm}
\bibfield{author}{\bibinfo{person}{Guolin Ke}, \bibinfo{person}{Qi Meng},
  \bibinfo{person}{Thomas Finley}, \bibinfo{person}{Taifeng Wang},
  \bibinfo{person}{Wei Chen}, \bibinfo{person}{Weidong Ma},
  \bibinfo{person}{Qiwei Ye}, {and} \bibinfo{person}{Tie-Yan Liu}.}
  \bibinfo{year}{2017}\natexlab{}.
\newblock \showarticletitle{Lightgbm: A highly efficient gradient boosting
  decision tree}. In \bibinfo{booktitle}{\emph{NeurIPS}}.
  \bibinfo{pages}{3146--3154}.
\newblock


\bibitem[\protect\citeauthoryear{Kingma and Ba}{Kingma and Ba}{2015}]%
        {adam2015method}
\bibfield{author}{\bibinfo{person}{Diederik~P Kingma} {and}
  \bibinfo{person}{Jimmy Ba}.} \bibinfo{year}{2015}\natexlab{}.
\newblock \showarticletitle{Adam: A method for stochastic optimization}. In
  \bibinfo{booktitle}{\emph{ICLR}}, Vol.~\bibinfo{volume}{5}.
\newblock


\bibitem[\protect\citeauthoryear{Li, Yosinski, Clune, Lipson, and Hopcroft}{Li
  et~al\mbox{.}}{2016}]%
        {visual2016convergent}
\bibfield{author}{\bibinfo{person}{Yixuan Li}, \bibinfo{person}{Jason
  Yosinski}, \bibinfo{person}{Jeff Clune}, \bibinfo{person}{Hod Lipson}, {and}
  \bibinfo{person}{John~E Hopcroft}.} \bibinfo{year}{2016}\natexlab{}.
\newblock \showarticletitle{Convergent learning: Do different neural networks
  learn the same representations?}. In \bibinfo{booktitle}{\emph{ICLR}}.
\newblock


\bibitem[\protect\citeauthoryear{Liu, Johns, and Davison}{Liu
  et~al\mbox{.}}{2019}]%
        {attention12019end}
\bibfield{author}{\bibinfo{person}{Shikun Liu}, \bibinfo{person}{Edward Johns},
  {and} \bibinfo{person}{Andrew~J Davison}.} \bibinfo{year}{2019}\natexlab{}.
\newblock \showarticletitle{End-to-end multi-task learning with attention}. In
  \bibinfo{booktitle}{\emph{CVPR}}. \bibinfo{pages}{1871--1880}.
\newblock


\bibitem[\protect\citeauthoryear{Long, Cao, Wang, and Philip}{Long
  et~al\mbox{.}}{2017}]%
        {tensor2017priors}
\bibfield{author}{\bibinfo{person}{Mingsheng Long}, \bibinfo{person}{Zhangjie
  Cao}, \bibinfo{person}{Jianmin Wang}, {and} \bibinfo{person}{S~Yu Philip}.}
  \bibinfo{year}{2017}\natexlab{}.
\newblock \showarticletitle{Learning multiple tasks with multilinear
  relationship networks}. In \bibinfo{booktitle}{\emph{NeurIPS}}.
  \bibinfo{pages}{1594--1603}.
\newblock


\bibitem[\protect\citeauthoryear{Ma, Zhao, Yi, Chen, Hong, and Chi}{Ma
  et~al\mbox{.}}{2018b}]%
        {mmoe2018kdd}
\bibfield{author}{\bibinfo{person}{Jiaqi Ma}, \bibinfo{person}{Zhe Zhao},
  \bibinfo{person}{Xinyang Yi}, \bibinfo{person}{Jilin Chen},
  \bibinfo{person}{Lichan Hong}, {and} \bibinfo{person}{Ed~H Chi}.}
  \bibinfo{year}{2018}\natexlab{b}.
\newblock \showarticletitle{Modeling task relationships in multi-task learning
  with multi-gate mixture-of-experts}. In \bibinfo{booktitle}{\emph{KDD}}.
  \bibinfo{pages}{1930--1939}.
\newblock


\bibitem[\protect\citeauthoryear{Ma, Zhao, Huang, Wang, Hu, Zhu, and Gai}{Ma
  et~al\mbox{.}}{2018a}]%
        {esmm2018sigir}
\bibfield{author}{\bibinfo{person}{Xiao Ma}, \bibinfo{person}{Liqin Zhao},
  \bibinfo{person}{Guan Huang}, \bibinfo{person}{Zhi Wang},
  \bibinfo{person}{Zelin Hu}, \bibinfo{person}{Xiaoqiang Zhu}, {and}
  \bibinfo{person}{Kun Gai}.} \bibinfo{year}{2018}\natexlab{a}.
\newblock \showarticletitle{Entire space multi-task model: An effective
  approach for estimating post-click conversion rate}. In
  \bibinfo{booktitle}{\emph{SIGIR}}. \bibinfo{pages}{1137--1140}.
\newblock


\bibitem[\protect\citeauthoryear{Qin, Cheng, Zhao, Chen, Metzler, and Qin}{Qin
  et~al\mbox{.}}{2020}]%
        {mmoe2020seq}
\bibfield{author}{\bibinfo{person}{Zhen Qin}, \bibinfo{person}{Yicheng Cheng},
  \bibinfo{person}{Zhe Zhao}, \bibinfo{person}{Zhe Chen},
  \bibinfo{person}{Donald Metzler}, {and} \bibinfo{person}{Jingzheng Qin}.}
  \bibinfo{year}{2020}\natexlab{}.
\newblock \showarticletitle{Multitask Mixture of Sequential Experts for User
  Activity Streams}. In \bibinfo{booktitle}{\emph{KDD}}.
  \bibinfo{pages}{3083--3091}.
\newblock


\bibitem[\protect\citeauthoryear{Ruder}{Ruder}{2017}]%
        {survey2017overview}
\bibfield{author}{\bibinfo{person}{Sebastian Ruder}.}
  \bibinfo{year}{2017}\natexlab{}.
\newblock \showarticletitle{An overview of multi-task learning in deep neural
  networks}.
\newblock \bibinfo{journal}{\emph{arXiv preprint arXiv:1706.05098}}
  (\bibinfo{year}{2017}).
\newblock


\bibitem[\protect\citeauthoryear{Shazeer, Mirhoseini, Maziarz, Davis, Le,
  Hinton, and Dean}{Shazeer et~al\mbox{.}}{2017}]%
        {moe32017outrageously}
\bibfield{author}{\bibinfo{person}{Noam Shazeer}, \bibinfo{person}{Azalia
  Mirhoseini}, \bibinfo{person}{Krzysztof Maziarz}, \bibinfo{person}{Andy
  Davis}, \bibinfo{person}{Quoc Le}, \bibinfo{person}{Geoffrey Hinton}, {and}
  \bibinfo{person}{Jeff Dean}.} \bibinfo{year}{2017}\natexlab{}.
\newblock \showarticletitle{Outrageously large neural networks: The
  sparsely-gated mixture-of-experts layer}. In
  \bibinfo{booktitle}{\emph{ICLR}}.
\newblock


\bibitem[\protect\citeauthoryear{Tang, Liu, Zhao, and Gong}{Tang
  et~al\mbox{.}}{2020}]%
        {cgc2020recsys}
\bibfield{author}{\bibinfo{person}{Hongyan Tang}, \bibinfo{person}{Junning
  Liu}, \bibinfo{person}{Ming Zhao}, {and} \bibinfo{person}{Xudong Gong}.}
  \bibinfo{year}{2020}\natexlab{}.
\newblock \showarticletitle{Progressive Layered Extraction (PLE): A Novel
  Multi-Task Learning (MTL) Model for Personalized Recommendations}. In
  \bibinfo{booktitle}{\emph{RecSys}}. \bibinfo{pages}{269--278}.
\newblock


\bibitem[\protect\citeauthoryear{Tzeng, Hoffman, Zhang, Saenko, and
  Darrell}{Tzeng et~al\mbox{.}}{2014}]%
        {mmd2014deep}
\bibfield{author}{\bibinfo{person}{Eric Tzeng}, \bibinfo{person}{Judy Hoffman},
  \bibinfo{person}{Ning Zhang}, \bibinfo{person}{Kate Saenko}, {and}
  \bibinfo{person}{Trevor Darrell}.} \bibinfo{year}{2014}\natexlab{}.
\newblock \showarticletitle{Deep domain confusion: Maximizing for domain
  invariance}.
\newblock \bibinfo{journal}{\emph{arXiv preprint arXiv:1412.3474}}
  (\bibinfo{year}{2014}).
\newblock


\bibitem[\protect\citeauthoryear{van~der Maaten and Hinton}{van~der Maaten and
  Hinton}{2008}]%
        {tsne2008}
\bibfield{author}{\bibinfo{person}{Laurens van~der Maaten} {and}
  \bibinfo{person}{Geoffrey Hinton}.} \bibinfo{year}{2008}\natexlab{}.
\newblock \showarticletitle{Visualizing Data using t-SNE}.
\newblock \bibinfo{journal}{\emph{Journal of Machine Learning Research}}
  \bibinfo{volume}{9}, \bibinfo{number}{86} (\bibinfo{year}{2008}),
  \bibinfo{pages}{2579--2605}.
\newblock


\bibitem[\protect\citeauthoryear{Vaswani, Shazeer, Parmar, Uszkoreit, Jones,
  Gomez, Kaiser, and Polosukhin}{Vaswani et~al\mbox{.}}{2017}]%
        {self2017attention}
\bibfield{author}{\bibinfo{person}{Ashish Vaswani}, \bibinfo{person}{Noam
  Shazeer}, \bibinfo{person}{Niki Parmar}, \bibinfo{person}{Jakob Uszkoreit},
  \bibinfo{person}{Llion Jones}, \bibinfo{person}{Aidan~N Gomez},
  \bibinfo{person}{Lukasz Kaiser}, {and} \bibinfo{person}{Illia Polosukhin}.}
  \bibinfo{year}{2017}\natexlab{}.
\newblock \showarticletitle{Attention is all you need}. In
  \bibinfo{booktitle}{\emph{NeurIPS}}. \bibinfo{pages}{5998--6008}.
\newblock


\bibitem[\protect\citeauthoryear{Wen, Zhang, Wang, Lv, Bao, Lin, and Yang}{Wen
  et~al\mbox{.}}{2020}]%
        {esm22020sigir}
\bibfield{author}{\bibinfo{person}{Hong Wen}, \bibinfo{person}{Jing Zhang},
  \bibinfo{person}{Yuan Wang}, \bibinfo{person}{Fuyu Lv},
  \bibinfo{person}{Wentian Bao}, \bibinfo{person}{Quan Lin}, {and}
  \bibinfo{person}{Keping Yang}.} \bibinfo{year}{2020}\natexlab{}.
\newblock \showarticletitle{Entire Space Multi-Task Modeling via Post-Click
  Behavior Decomposition for Conversion Rate Prediction}. In
  \bibinfo{booktitle}{\emph{SIGIR}}. \bibinfo{pages}{2377--2386}.
\newblock


\bibitem[\protect\citeauthoryear{Xi, Song, Zhuang, Zhu, Chen, Zhang, Qi, and
  He}{Xi et~al\mbox{.}}{2021}]%
        {difm2021modeling}
\bibfield{author}{\bibinfo{person}{Dongbo Xi}, \bibinfo{person}{Bowen Song},
  \bibinfo{person}{Fuzhen Zhuang}, \bibinfo{person}{Yongchun Zhu},
  \bibinfo{person}{Shuai Chen}, \bibinfo{person}{Tianyi Zhang},
  \bibinfo{person}{Yuan Qi}, {and} \bibinfo{person}{Qing He}.}
  \bibinfo{year}{2021}\natexlab{}.
\newblock \showarticletitle{Modeling the Field Value Variations and Field
  Interactions Simultaneously for Fraud Detection}. In
  \bibinfo{booktitle}{\emph{AAAI}}.
\newblock


\bibitem[\protect\citeauthoryear{Xi, Zhuang, Liu, Gu, Xiong, and He}{Xi
  et~al\mbox{.}}{2019}]%
        {bistddp2019modelling}
\bibfield{author}{\bibinfo{person}{Dongbo Xi}, \bibinfo{person}{Fuzhen Zhuang},
  \bibinfo{person}{Yanchi Liu}, \bibinfo{person}{Jingjing Gu},
  \bibinfo{person}{Hui Xiong}, {and} \bibinfo{person}{Qing He}.}
  \bibinfo{year}{2019}\natexlab{}.
\newblock \showarticletitle{Modelling of Bi-Directional Spatio-Temporal
  Dependence and Users’ Dynamic Preferences for Missing POI Check-In
  Identification}. In \bibinfo{booktitle}{\emph{AAAI}},
  Vol.~\bibinfo{volume}{33}. \bibinfo{pages}{5458--5465}.
\newblock


\bibitem[\protect\citeauthoryear{Xi, Zhuang, Song, Zhu, Chen, Hong, Chen, Gu,
  and He}{Xi et~al\mbox{.}}{2020}]%
        {nhfm2020neural}
\bibfield{author}{\bibinfo{person}{Dongbo Xi}, \bibinfo{person}{Fuzhen Zhuang},
  \bibinfo{person}{Bowen Song}, \bibinfo{person}{Yongchun Zhu},
  \bibinfo{person}{Shuai Chen}, \bibinfo{person}{Dan Hong},
  \bibinfo{person}{Tao Chen}, \bibinfo{person}{Xi Gu}, {and}
  \bibinfo{person}{Qing He}.} \bibinfo{year}{2020}\natexlab{}.
\newblock \showarticletitle{Neural Hierarchical Factorization Machines for
  User's Event Sequence Analysis}. In \bibinfo{booktitle}{\emph{SIGIR}}.
  \bibinfo{pages}{1893--1896}.
\newblock


\bibitem[\protect\citeauthoryear{Xiao, Ye, He, Zhang, Wu, and Chua}{Xiao
  et~al\mbox{.}}{2017}]%
        {xiao2017afm}
\bibfield{author}{\bibinfo{person}{Jun Xiao}, \bibinfo{person}{Hao Ye},
  \bibinfo{person}{Xiangnan He}, \bibinfo{person}{Hanwang Zhang},
  \bibinfo{person}{Fei Wu}, {and} \bibinfo{person}{Tat-Seng Chua}.}
  \bibinfo{year}{2017}\natexlab{}.
\newblock \showarticletitle{Attentional factorization machines: Learning the
  weight of feature interactions via attention networks}. In
  \bibinfo{booktitle}{\emph{IJCAI}}.
\newblock


\bibitem[\protect\citeauthoryear{Yang and Hospedales}{Yang and
  Hospedales}{2017}]%
        {tensor2017deep}
\bibfield{author}{\bibinfo{person}{Yongxin Yang} {and} \bibinfo{person}{Timothy
  Hospedales}.} \bibinfo{year}{2017}\natexlab{}.
\newblock \showarticletitle{Deep multi-task representation learning: A tensor
  factorisation approach}. In \bibinfo{booktitle}{\emph{ICLR}}.
\newblock


\bibitem[\protect\citeauthoryear{Zeiler and Fergus}{Zeiler and Fergus}{2014}]%
        {visual2014visualizing}
\bibfield{author}{\bibinfo{person}{Matthew~D Zeiler} {and} \bibinfo{person}{Rob
  Fergus}.} \bibinfo{year}{2014}\natexlab{}.
\newblock \showarticletitle{Visualizing and understanding convolutional
  networks}. In \bibinfo{booktitle}{\emph{ECCV}}. \bibinfo{pages}{818--833}.
\newblock


\bibitem[\protect\citeauthoryear{Zhao, Du, Sun, Zhuang, Lv, and Xiong}{Zhao
  et~al\mbox{.}}{2019a}]%
        {attention2019z}
\bibfield{author}{\bibinfo{person}{Jiejie Zhao}, \bibinfo{person}{Bowen Du},
  \bibinfo{person}{Leilei Sun}, \bibinfo{person}{Fuzhen Zhuang},
  \bibinfo{person}{Weifeng Lv}, {and} \bibinfo{person}{Hui Xiong}.}
  \bibinfo{year}{2019}\natexlab{a}.
\newblock \showarticletitle{Multiple Relational Attention Network for
  Multi-task Learning}. In \bibinfo{booktitle}{\emph{KDD}}.
  \bibinfo{pages}{1123--1131}.
\newblock


\bibitem[\protect\citeauthoryear{Zhao, Hong, Wei, Chen, Nath, Andrews,
  Kumthekar, Sathiamoorthy, Yi, and Chi}{Zhao et~al\mbox{.}}{2019b}]%
        {mmo22019video}
\bibfield{author}{\bibinfo{person}{Zhe Zhao}, \bibinfo{person}{Lichan Hong},
  \bibinfo{person}{Li Wei}, \bibinfo{person}{Jilin Chen},
  \bibinfo{person}{Aniruddh Nath}, \bibinfo{person}{Shawn Andrews},
  \bibinfo{person}{Aditee Kumthekar}, \bibinfo{person}{Maheswaran
  Sathiamoorthy}, \bibinfo{person}{Xinyang Yi}, {and} \bibinfo{person}{Ed
  Chi}.} \bibinfo{year}{2019}\natexlab{b}.
\newblock \showarticletitle{Recommending what video to watch next: a multitask
  ranking system}. In \bibinfo{booktitle}{\emph{RecSys}}.
  \bibinfo{pages}{43--51}.
\newblock


\bibitem[\protect\citeauthoryear{Zhu, Xi, Song, Zhuang, Chen, Gu, and He}{Zhu
  et~al\mbox{.}}{2020}]%
        {www2020modeling}
\bibfield{author}{\bibinfo{person}{Yongchun Zhu}, \bibinfo{person}{Dongbo Xi},
  \bibinfo{person}{Bowen Song}, \bibinfo{person}{Fuzhen Zhuang},
  \bibinfo{person}{Shuai Chen}, \bibinfo{person}{Xi Gu}, {and}
  \bibinfo{person}{Qing He}.} \bibinfo{year}{2020}\natexlab{}.
\newblock \showarticletitle{Modeling Users’ Behavior Sequences with
  Hierarchical Explainable Network for Cross-domain Fraud Detection}. In
  \bibinfo{booktitle}{\emph{TheWebConf}}. \bibinfo{pages}{928--938}.
\newblock


\end{thebibliography}

\appendix
\section{Appendix}
\begin{table}[!t]
  \centering
  \caption{The summary of the hyper-parameters in multi-task models, which includes ESMM, OMoE, MMoE, PLE and \model. The $T$ is the number of tasks. Except for these listed, other parts of these models that involve MLP are all single-layer.}
    \begin{tabular}{cc}
    \toprule
    Hyper-parameter & Value \\
    \midrule
    Optimizer & Adam \\
    Batch size & 2000 \\
    Learning rate & 1e-3 \\
    L2 regularization & 1e-6 \\
    Embedding dimension & 5 \\
    Dimensions of layers in the MLP-& [64,32,16]$\times$2$\times T$\\ Expert of Expert-Bottom pattern & \\
    Dimensions of layers in the MLP-& [128,64,32]$\times T$\\Tower of Probability-Transfer pattern & \\
    Dropout rate in each layers & [0.1,0.3,0.3] \\
    Activation function in MLP & Relu \\
    \bottomrule
    \end{tabular}%
  \label{tab:params}%
\end{table}%
\subsection{Reproducibility Information}
For the LightGBM model, the learning rate, feature fraction, bagging fraction, bagging frequency, max bin, number of leaves, boosting type are set as
0.1, 0.9, 0.7, 5, 2000, 70, gbdt, respectively, 
which are chosen according to the grid search on the validation set. 

For the neural network-based models, to verify the generalization ability of different models and compare them fairly, different neural network-based models use the same common hyper-parameters on two datasets considering the empirical values and computational efficiency.
For the industrial dataset, we downsample the \activate~ negative samples to keep the proportion $\lambda$ of positive samples at $1\%$ except for the test set.
For both industrial and public datasets, we use: embedding dimension $d=5$, the output dimension of the Tower is $k=32$, the strength of the \constraint~ is $\alpha=0.6$. 
We only fine-tune the hyper-parameters of $d$, $\alpha$, and $\lambda$ according to grid search on the validation set. 
Besides, the Tower $f_t(\cdot)$ is a three-layer MLP with dimension $[128,64,32]$, the $g_t(\cdot)$, $h_1(\cdot)$, $h_2(\cdot)$ and $h_3(\cdot)$ are all single-layer MLP with dimension $32$. 
For a fair comparison, we try our best to ensure that the main architecture of different multi-task models is consistent. The summary of different multi-task models is shown in Table \ref{tab:params}.
We conduct experiments of all models with NVIDIA Tesla V100 GPU with 16G memory.

\subsection{More Experiments}
These inspiring observations, which are described in Subsection \ref{sub:ablation} and shown in Figure \ref{fig:activate} on the \activate~ task, can also be observed on the \credit~ task.
Firstly, we randomly sample $500$ \credit~ positive and negative samples in the test set, respectively.
The \credit~ prediction scores of positive samples are ranked in descending order, while those of negative samples are in ascending order.
Then, we plot the original information $\q_t$, transferred plus original information $\p_{t-1}+\q_t$ and the information $\z_t$ learned by the \module~ on the \credit~ task via the t-SNE.
We show the results in Figure \ref{fig:credit}.

\subsection{Data Collection and Privacy Protection}
The industrial dataset contains all samples that are shown a banner of Meituan Co-Branded Credit Cards over a continuous period of time. 
In the traditional credit card business, the \credit~ step is usually not real-time. However, in our online credit card application, Meituan and the card-issuing bank make two-level real-time risk judgment, which makes the \credit~ step is almost real-time. Besides, the \activate~ step requires the user to have received the mailed credit card and go to the bank to activate it or make an appointment with a salesman to activate it at home.
Therefore, the samples used are shown a banner at least 14 days ago, so as to ensure the accuracy of the \activate~ label.
The features used include context features and user statistics features.
We don't use any features that can locate specific users and involve user privacy.
The accurate end-to-end conversion identification can disturb users as little as possible and improve the user experience.

\end{document}